\documentclass[conference]{IEEEtran}
\usepackage{times}
\usepackage{amsmath}

\usepackage{multicol}
\usepackage{graphicx}
\usepackage{tikz}
\usepackage{flafter}  

\usepackage[
    colorinlistoftodos,
    prependcaption,
    textsize=tiny,
    obeyFinal,
]{todonotes}
\usepackage[ruled]{algorithm2e}
\SetKwComment{Comment}{\# }{}  
\SetArgSty{textnormal}

\usepackage[hidelinks]{hyperref}
\usepackage[
    nameinlink,  
    noabbrev,  
    capitalize,  
]{cleveref}  

\usepackage[url=false, sorting=none, maxbibnames=6]{biblatex}
\usepackage{fancyhdr}
\fancypagestyle{withfooter}{
  
  \fancyfoot[C]{\footnotesize Accepted to the RSS Workshop on Semantics for Robotics: From Environment Understanding and Reasoning to Safe Interaction 2024}
}

\newcounter{symbolCounter}
\newcommand{\getThanksSymbol}[1]{%
\setcounter{symbolCounter}{#1}%
$^\fnsymbol{symbolCounter}$%
}

\newcommand{\makeThanksSection}
{%
\renewcommand{\thefootnote}{\fnsymbol{footnote}}%
\footnotetext[1]{TNO, \textit{Autonomous Systems \& Decision Support}, Netherlands}%
\footnotetext[2]{Delft University of Technology, \textit{Signal Processing Systems}, Netherlands}%
\renewcommand{\thefootnote}{\arabic{footnote}}%
}

\title{Scaling 3D Reasoning with LMMs to Large Robot Mission Environments Using Datagraphs}

\author{
W.J. Meijer\getThanksSymbol{1} \and 
A.C. Kemmeren\getThanksSymbol{1} \and 
E.H.J. Riemens\getThanksSymbol{2} \and 
J.E. Fransman\getThanksSymbol{1} \and 
M. van Bekkum\getThanksSymbol{1} \and 
G.J. Burghouts\getThanksSymbol{1} \and 
J.D. van Mil\getThanksSymbol{1}
}

\begin{document}
\maketitle
\makeThanksSection

\thispagestyle{withfooter}
\pagestyle{withfooter}

\begin{abstract}
This paper addresses the challenge of scaling Large Multimodal Models (LMMs) to expansive 3D environments. 
Solving this open problem is especially relevant for robot deployment in many first-responder scenarios, such as search-and-rescue missions that cover vast spaces.
The use of LMMs in these settings is currently hampered by the strict context windows that limit the LMM's input size.  
We therefore introduce a novel approach that utilizes a datagraph structure, which allows the LMM to iteratively query smaller sections of a large environment. 
Using the datagraph in conjunction with graph traversal algorithms, we can prioritize the most relevant locations to the query, thereby improving the scalability of 3D scene language tasks. 
We illustrate the datagraph using 3D scenes, but these can be easily substituted by other dense modalities that represent the environment, such as pointclouds or Gaussian splats.  
We demonstrate the potential to use the datagraph for two 3D scene language task use cases, in a search-and-rescue mission example.
\end{abstract}

\IEEEpeerreviewmaketitle

\section{Introduction}

\begin{figure*}[t!]
    \centering
    \includegraphics[trim=0 300 0 0, clip,width=0.75\linewidth]{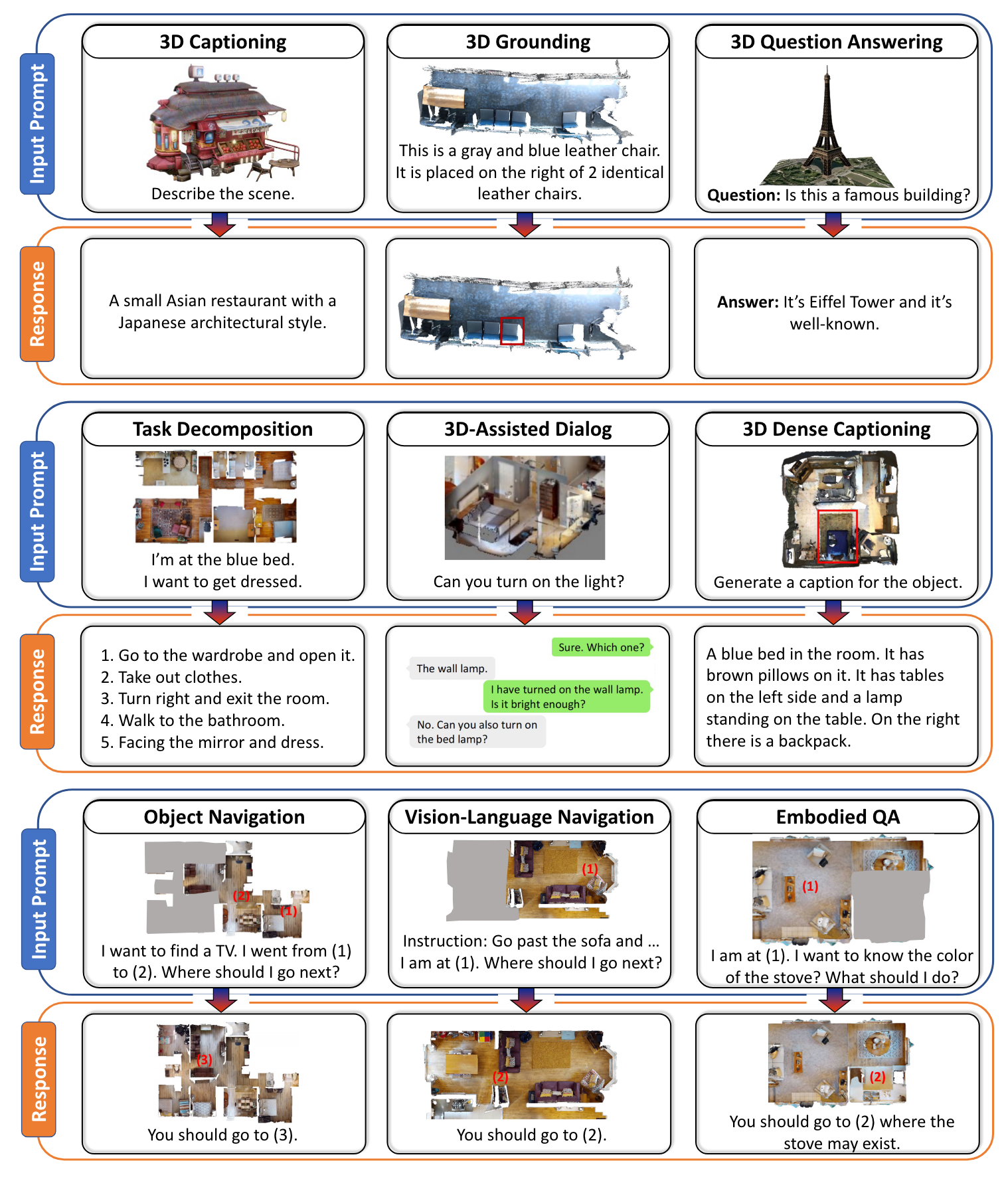}
    \caption{There is a variety of 3D tasks we would like to perform with 3D-LMMs as illustrated in the work of \textcite{hong_3d-llm_2023}.
    We investigate how to perform such tasks over expansive environments with existing LMMs, which are limited in their context size. } 
    \label{fig:3Dtasks}
\end{figure*}

This paper addresses the challenge of scaling the use of Large Multi-modal Models (LMMs) to large 3D environments. 
Current models can perform a wide range of 3D scene language tasks such as spatial and semantic reasoning (\cref{fig:3Dtasks}, \textcite{hong_3d-llm_2023}), but the impressive capabilities of these models do not yet transfer to large environments that are generally encountered in search-and-rescue settings.

In these large environments, current 3D LMMs seem to face a similar limitation as many Vision Language Models (VLMs), namely the \emph{resolution curse problem}\footnote{\url{https://huggingface.co/blog/visheratin/vlm-resolution-curse}}. 
The VLMs cannot properly encode the information in high-resolution input images due to the fixed and limited input size of the model \cite{han_survey_2023}. 
Similarly, the context window size in transformer-based models limits the maximum number of tokens that can be used \cite{dai_transformer-xl_2019, sanford_representational_nodate}. This limits current 3D LMMs to be able to process only small areas. For example, current models have not yet been used on areas larger than 10 rooms \cite{hong_3d-llm_2023, chen_leveraging_2023}.

Recent developments tackle this issue by enabling transformer-based models to process contexts of infinite length flexibly \cite{munkhdalai_leave_2024}. 
However, it will take substantial effort to develop the models that are capable of performing all 3D scene tasks in \cref{fig:3Dtasks} in large 3D environments. 
These developments are slowed by a lack of large 3D scene datasets \cite{jia_sceneverse_2024} and the vast hardware requirements for training new LMMs \cite{munkhdalai_leave_2024}. 
In the meantime, the approach in this paper will enable us to use \textit{existing} 3D LMMs in large environments instead.

\begin{figure}
    \centering
    \includegraphics[width=\columnwidth]{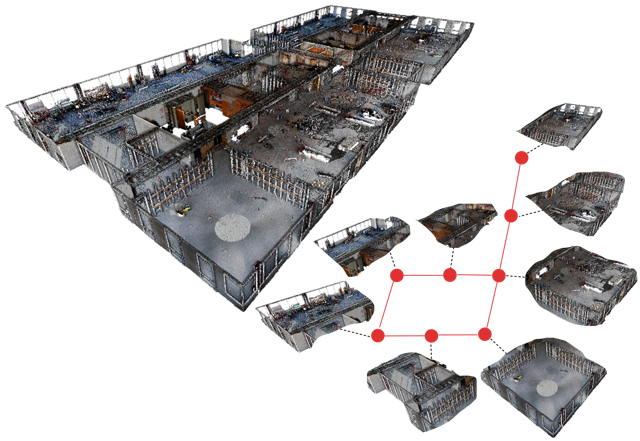}
    \caption{On the left is a continuous multi-room 3D scene. On the right, a graph (red) is extended with 3D scenes at each node to form a datagraph. Instead of processing the whole multi-room 3D scene at once, existing LMMs can use the smaller scenes in the datagraph to iteratively cover large areas. }
    \label{fig:data graph}
\end{figure}

Our approach is inspired by how the resolution curse is often solved for 2D tasks, using \emph{tiling}. 
In tiling, the input image is subsampled into several tiles of lower resolution.
In more advanced works, strategies to dynamically choose these tiles are proposed, by internal dialogue with a Large Language Model and visual memory \cite{wu_v_2023} or by exploiting the attention of the transformer \cite{zhang_towards_2024}.
Similarly, we divide large 3D environments into smaller areas by exploiting spatial structure and then choose the subset of areas that is potentially relevant to the task (\cref{fig:data graph}). 
This, for example, enables us to only consider the part of the environment that is near and reachable to a robot.

In this paper, we consider a scene representation that retains full 3D information, because we believe that having access to detailed information is crucial to search-and-rescue missions. 
For example, there is an important distinction between a person lying on the floor of the hallway and a person lying in bed.
This is often not captured when using 3D scene graphs as representation, as context information is lost during the compression steps in the mapping procedure \cite{armeni_3d_2019, rosinol_3d_2020}. 
Subsequent reasoning tasks then cannot access important contextual details that were initially captured by the sensors \cite{rana_sayplan_2023}. 

We propose to leverage a spatial graph structure to subdivide vast 3D scenes into smaller areas (\cref{fig:data graph}). 
With existing graph traversal algorithms, smaller chunks of 3D information can be fed iteratively into the LMMs.
We demonstrate the potential of this approach using two algorithms. 
\cref{alg:search} starts querying locations closer to the robot, which is useful, for example, when searching for an object nearby.
Alternatively, \cref{alg:route} queries along a path, for example enabling an LMM to check for situations that make navigation along that path hazardous. 
These approaches bypass the context limitations of current LMMs, while allowing them to scale to vast environments.
Moreover, the approach can be easily extended to store other types of information about the environment. 
The 3D scenes can be substituted with other modalities, e.g. when deploying LMMs that work with \emph{(colored) pointclouds} \cite{xu_pointllm_2023, qi_gpt4point_2023}, \emph{neural radiance fields} \cite{kerr_lerf_2023, bruns_neural_2024}, or \emph{Gaussian splats} \cite{qin_langsplat_2024}.




\section{Approach}
This work considers robot deployments for time-critical first-responder scenarios, such as search-and-rescue missions.
First responders operate in expansive and unknown environments, where it is important to quickly build situational awareness (SA) \cite{sanquist_attention_nodate}, and reason about complex questions such as the safety of an environment. 
For this, we use currently existing LMMs and combine them with a datagraph to bypass the LMM's limited context size. The following sections provide more details on the datagraph and how existing graph traversal algorithms then allow for iterative prompting of the LMM.

%
\subsection{The datagraph}
The datagraph is formalized as $G = (V, E)$ with node set $V$ and edge set $E$.
Each node $v \in V$ has a pose $x$ and a data snapshot $s$, which makes the tuple $v = (x_v, s_v)$.
An edge $e_{ij} = (v_i, v_j)$ connects two nodes for neighboring areas. 
One could consider additional requirements that an edge is only added if there is a traversable route between the two areas \cite{meijer_situational_2024}.
This would make $G$ compatible with navigation or exploration graphs commonly used in autonomy stacks \cite{hudson_heterogeneous_2022, meijer_situational_2024, tranzatto_cerberus_2022}.
The $s$ in the node can be data of any modality that can be used to query an LMM, including 3D scenes, Gaussian splats, or images. 

\subsection{Graph traversal for iterative LMM prompting}
Depending on the task that the LMM will execute, different parts of the environment could be interesting to prompt. The challenge for graph traversal thus is to prompt areas with highest priority first. We exemplify this with two graph traversal algorithms. The first algorithm (\cref{alg:search}) prioritizes the areas that are closest to the robot. 
The second algorithm (\cref{alg:route}) retrieves information along a path through the environment to enable, for example, safer navigation.
\\

\begin{figure}
    \centering
    \includegraphics[width=\columnwidth]{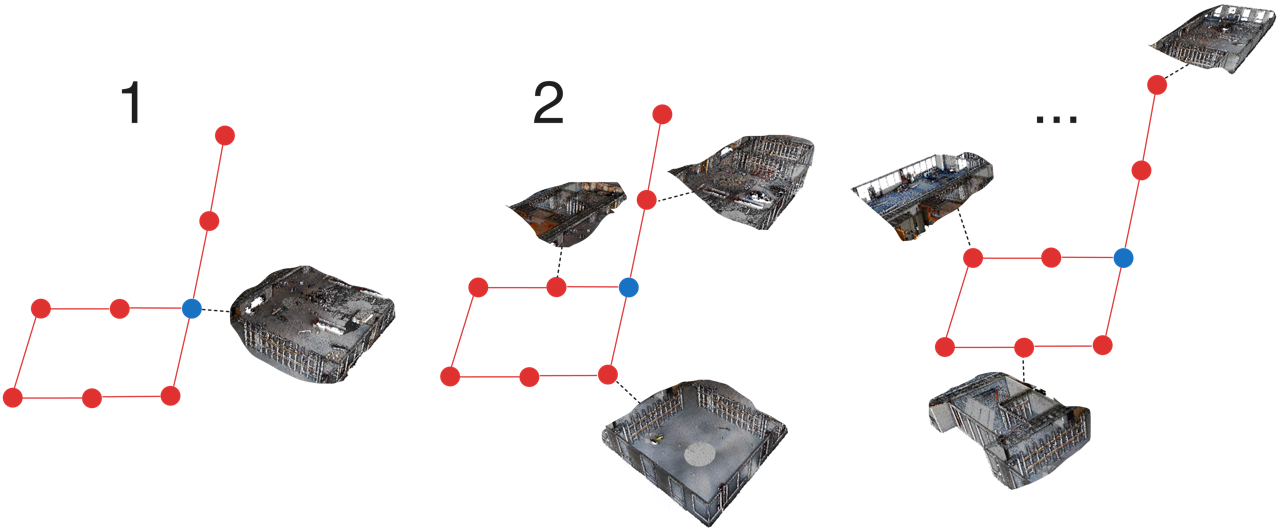}
    \caption{Illustration of the iterative spatially grounded LMM querying algorithm. Search is started at the node the agent is at (blue), then it gradually expands outward to retrieve perception scenes that can be queried by the LMM. }
    \label{fig:search}
\end{figure}

\subsubsection{3D scene language tasks prioritized by robot proximity}
Various situations may require a robot to first recall areas closest to its current situation and execute 3D scene tasks in these areas. 
A clear example in search-and-rescue missions is when searching for objects in the environment. 
A robot may need to go through a locked door and start to search in its memory (the datagraph) if it has seen tools that may help it to open or break this door. 
Finding the closest satisfactory tool then helps fulfilling the mission in the least amount of time. 
Another example of object search is when new information about risks in the environment becomes available during the mission, such as indicators of explosive or toxic materials. 
It is important for the agent to be aware of the locations of such materials, especially if the hazard is close. 

\cref{alg:search} starts with retrieving the nodes closest to the agent, in a breadth-first manner.
It begins by initializing an empty set $V_{visited}$ to track the visited nodes and a set $V_{search}$ containing the starting node $v_{agent}$.
For each node $v$ in $V_{search}$, it retrieves the scene associated with $v$ from the datagraph $G$ and queries the LMM with this scene and the query $q$. 
Then, it marks $v$ as visited by adding it to $V_{visited}$ and updates $V_{search}$ with the neighbors of $v$ that have not yet been visited. 
This process repeats until the entire graph is traversed. The steps are summarized in \cref{alg:search}.

For specific tasks, one may consider altering this algorithm to stop when a satisfactory response is found. For example, there is little reason to continue searching for a tool to open a door after a suitable object is found. However, in the case of searching for hazards, one may naturally want to continue these queries even if the first hazardous area is found.
\begin{algorithm}
\caption{Iteratively query an LMM by expanding the datagraph spatially.}
\label{alg:search}
\KwData{datagraph $G$, query $q$, agent location $v_\mathrm{agent}$}
\KwResult{LLM responses $R$ for each node $v\in V$}
$V_\mathrm{visited} \gets \{\}$\;
$V_\mathrm{search} \gets \{v_\mathrm{agent}\}$\;
    \For{$v \; \textbf{in} \; V_\mathrm{search}$}{
        scene = $G[v]$\;
        response = LMM(scene, $q$)\;
        $R\text{.append(response)}$\;
        $V_\mathrm{visited} \gets V_\mathrm{visited} \cup v$\;
        $V_\mathrm{search} \gets $ \texttt{neighbors($v$)} \text{not} \textbf{in} $V_\mathrm{visited}$\;        
    }
    \Return $R$\;
\end{algorithm}
\begin{algorithm}
\caption{Query an LMM along a path.}
\label{alg:route}
\KwData{datagraph $G$, query $q$, path $P$}
\KwResult{LMM responses $R$ for each node in path $P$}

    \For{$v\; \textbf{in} \; P$}{
        scene = $G[v]$\;
        response = LMM(scene, $q$)\;
        $R\text{.append(response)}$ \;
    }
\Return $R$\;
\end{algorithm}

\subsubsection{3D scene language tasks along a navigation path}
Consider a search-and-rescue mission, where a robot guides a victim to safety. Assessing the safety of traversing routes through the environment is vital. 
Therefore, we may want to query whether there are any hazards along the proposed routes. 

\begin{figure}
    \centering
    \includegraphics[width=\columnwidth]{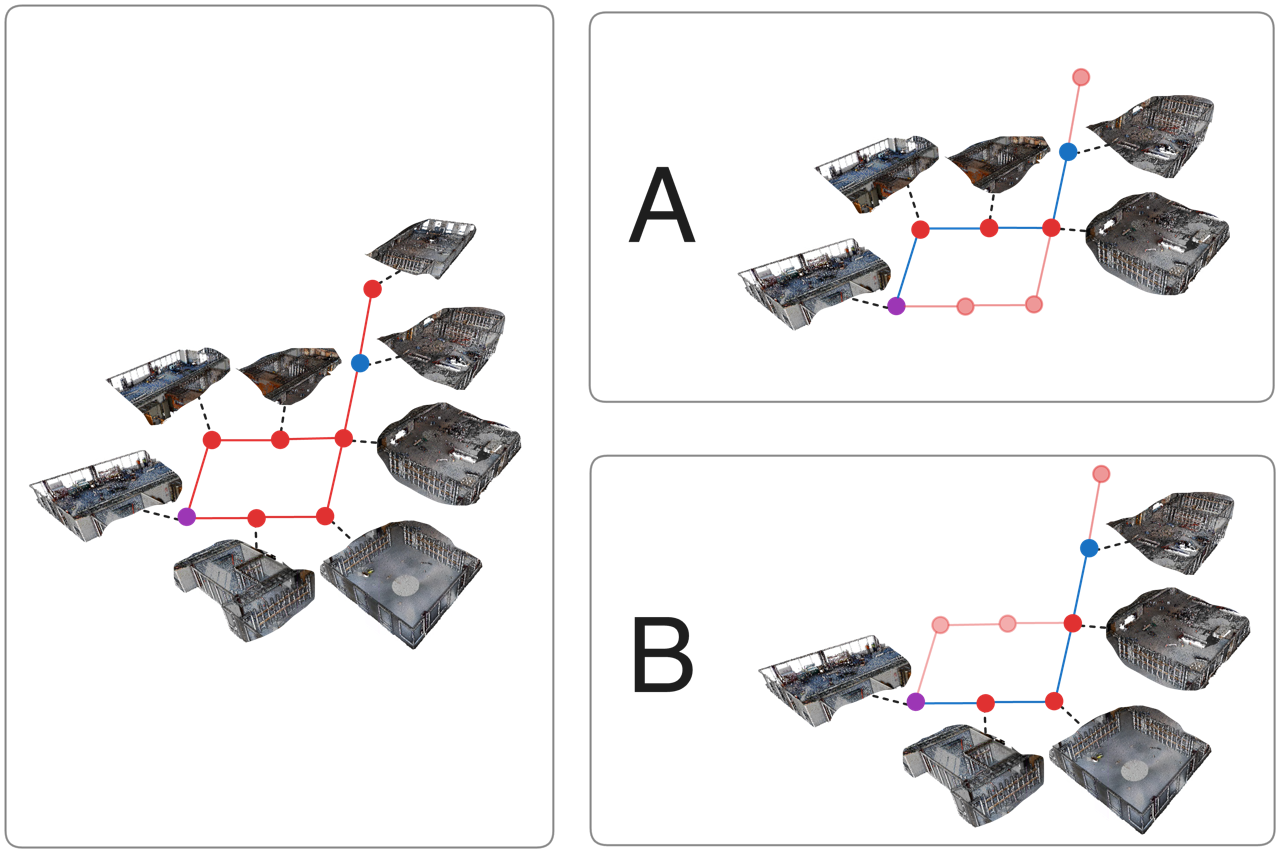}
    \caption{Illustration of 3D scene language tasks along a navigation path. The purple node is the navigation goal and the blue node is where the agent is located. Two routes A and B are possible, highlighted with blue edges in the two figures on the right.
    }
    \label{fig:routeDiagram}
\end{figure}

Given a path in the datagraph, we can iteratively query the LMM about the safety of an area using the data in each of the nodes, as illustrated in \cref{fig:routeDiagram}. 
The algorithm iterates through each node $v$, retrieves the scene associated with $v$ from $G$, and queries the LMM with this scene context and query $q$. 
Responses are added to the list $R$. 
The algorithm is summarized in \cref{alg:route}.\\

Because our method only considers local scenes, it will require additional logic to answer certain global questions. 
For example, when querying how many times a certain object is present in all the scenes of the environment, we would need to aggregate the individual answers per scene (local node) over all scenes of the environment (all relevant nodes).\\


\section{Discussion and Conclusion}

In this paper, we have introduced a novel datagraph structure that significantly expands the spatial context over which Large Multimodal Models (LMMs) can reason. 
The method was illustrated using 3D scenes, but in principle it is agnostic to the data type stored in the graph. 
The graph allows us to prioritize data from locations that are most relevant to the query, such as those closest to the agent. 
This prioritization improves the scalability of 3D scene language tasks, enabling their use in expansive robotic mission environments.

Looking ahead, there are several challenges and opportunities for future work. 
One such challenge arises at the boundaries between scenes. 
Relevant information might end up appearing in both scenes, for example, leading to duplicate detections, or scattered over distinct scenes, negatively affecting performance.
Moreover, we see a need for experimental results to support our proposed approach. 
For instance, stitching the rooms of the SceneVerse dataset together could form large environments that would provide a robust testbed for our approach. 
Additionally, investigating methods of caching identical queries along a path for Algorithm 2 could further enhance the efficiency and effectiveness of our approach.

In conclusion, our work presents a relevant step forward in the application of LMMs in large environments. 
By leveraging the datagraph structure and prioritizing relevant locations, we have demonstrated a scalable solution for 3D scene language tasks in large robotic mission environments. 
As we continue to refine our approach and address the challenges identified, we are optimistic about the potential impact of our work on the future of robotic missions.





\printbibliography

\end{document}